\documentclass[runningheads]{llncs}
\usepackage{graphicx}

\usepackage{eccv}
\usepackage{tikz}
\usepackage{comment}
\usepackage{amsmath,amssymb} 
\usepackage{color}
\usepackage{multirow}
\usepackage{graphicx}
\usepackage{amsmath}
\usepackage{amssymb}
\usepackage{boldline}
\usepackage{algorithm}
\usepackage{algorithmic}

\usepackage{multicol}
\usepackage{booktabs}
\usepackage{float}

\usepackage[accsupp]{axessibility}  

\usepackage[width=122mm,left=12mm,paperwidth=146mm,height=193mm,top=12mm,paperheight=217mm]{geometry}
\usepackage{hyperref}
\usepackage{microtype}
 
\begin{document}
\pagestyle{headings}
\mainmatter

\title{Stealing Stable Diffusion Prior for Robust Monocular Depth Estimation} 

\titlerunning{ } 
\authorrunning{ } 
\author{Yifan Mao\thanks{Equal contribution.}~~~ Jian Liu$^{*}$~~~Xianming Liu$\thanks{Corresponding Author.}$ \\ }
\institute{Harbin Institute of Technology}

\maketitle

\begin{abstract} 
 Monocular depth estimation is a crucial task in computer vision. While existing methods have shown impressive results under standard conditions, they often face challenges in reliably performing in scenarios such as low-light or rainy conditions due to the absence of diverse training data. This paper introduces a novel approach named Stealing Stable Diffusion (SSD) prior for robust monocular depth estimation. The approach addresses this limitation by utilizing stable diffusion to generate synthetic images that mimic challenging conditions. Additionally, a self-training mechanism is introduced to enhance the model's depth estimation capability in such challenging environments. To enhance the utilization of the stable diffusion prior further, the DINOv2 encoder is integrated into the depth model architecture, enabling the model to leverage rich semantic priors and improve its scene understanding. Furthermore, a teacher loss is introduced to guide the student models in acquiring meaningful knowledge independently, thus reducing their dependency on the teacher models. The effectiveness of the approach is evaluated on nuScenes and Oxford RobotCar, two challenging public datasets, with the results showing the efficacy of the method. Source code are available at: \href{https://github.com/hitcslj/SSD}{https://github.com/hitcslj/SSD}.

\end{abstract}

\section{Introduction}
\label{sec:intro}

Monocular depth estimation (MDE) is crucial in computer vision, providing important cues for various downstream applications such as autonomous driving and robotics. However, obtaining accurate 3D depth information from a single image presents a geometrically ill-posed challenge. Traditional methods like stereo matching and structure from motion have demonstrated restricted performance in this regard \cite{scharstein2002taxonomy, hartley2003multiple}. Since 2014, the emergence of deep learning has markedly enhanced depth estimation performance. Deep learning models can acquire rich prior knowledge from data, facilitating scene understanding and presenting promising solutions for depth estimation. Consequently, numerous methods for monocular depth estimation have emerged, encompassing supervised approaches \cite{NIPS2014_multiscale_depth, Eigen_2015_ICCV, 7346484, Adabins_Bhat_2021_CVPR, li2022binsformer, yuan2022newcrfs} and self-supervised methods \cite{10.1007/978-3-319-46484-8_45, Zhou_2017_CVPR, Godard_2017_CVPR, monodepth2}.

Although MDE methods perform well under standard conditions, such as sunny weather, they become less effective in challenging conditions like darkness and adverse weather. These limitations arise from invalidating crucial assumptions, such as photometric consistency and reliable ground truth, in challenging scenarios. Additionally, current datasets lack sufficient samples capturing challenging scenarios, with a notable scarcity of dedicated datasets tailored to address these conditions. Recent research has explored Robust Monocular Depth Estimation (RMDE) under challenging conditions, categorizing them into two groups: model-based and data-based approaches. Model-based approaches \cite{Wang_2021_ICCV,wsgd,10.1145/3581783.3611807,zhao2022monovit,spencer2020defeat} aim to enhance the model's capability of handling challenging conditions by modifying its architecture. Conversely, data-based approaches achieve RMDE by enhancing the image signal \cite{adfa,9803821,md4all,Saunders_2023_ICCV,liu2021self} through techniques such as domain adaptation or utilizing data from other modalities \cite{9340998,9665968,Shin_2023_CVPR}.

Although previous model-based methods have achieved satisfactory results, they rely on complex pipelines tailored for specific challenging conditions. This approach constrains their ability to reason and adapt to diverse challenging conditions. On the other hand, methods based on data from other modalities face challenges in obtaining high-quality data and often require post-processing. Augmenting the image signal in data-based methods can mitigate some of these limitations by employing simpler model architectures. These methods utilize GANs as translation models to generate training samples; however, GANs often lack generalizability and demonstrate limited diversity in the generated samples. Furthermore, adapting the model to multiple domains necessitates training multiple GANs, leading to additional training costs.

\begin{figure}[ht]
\vspace{-8mm}
\centering
\includegraphics[width=0.85\linewidth]{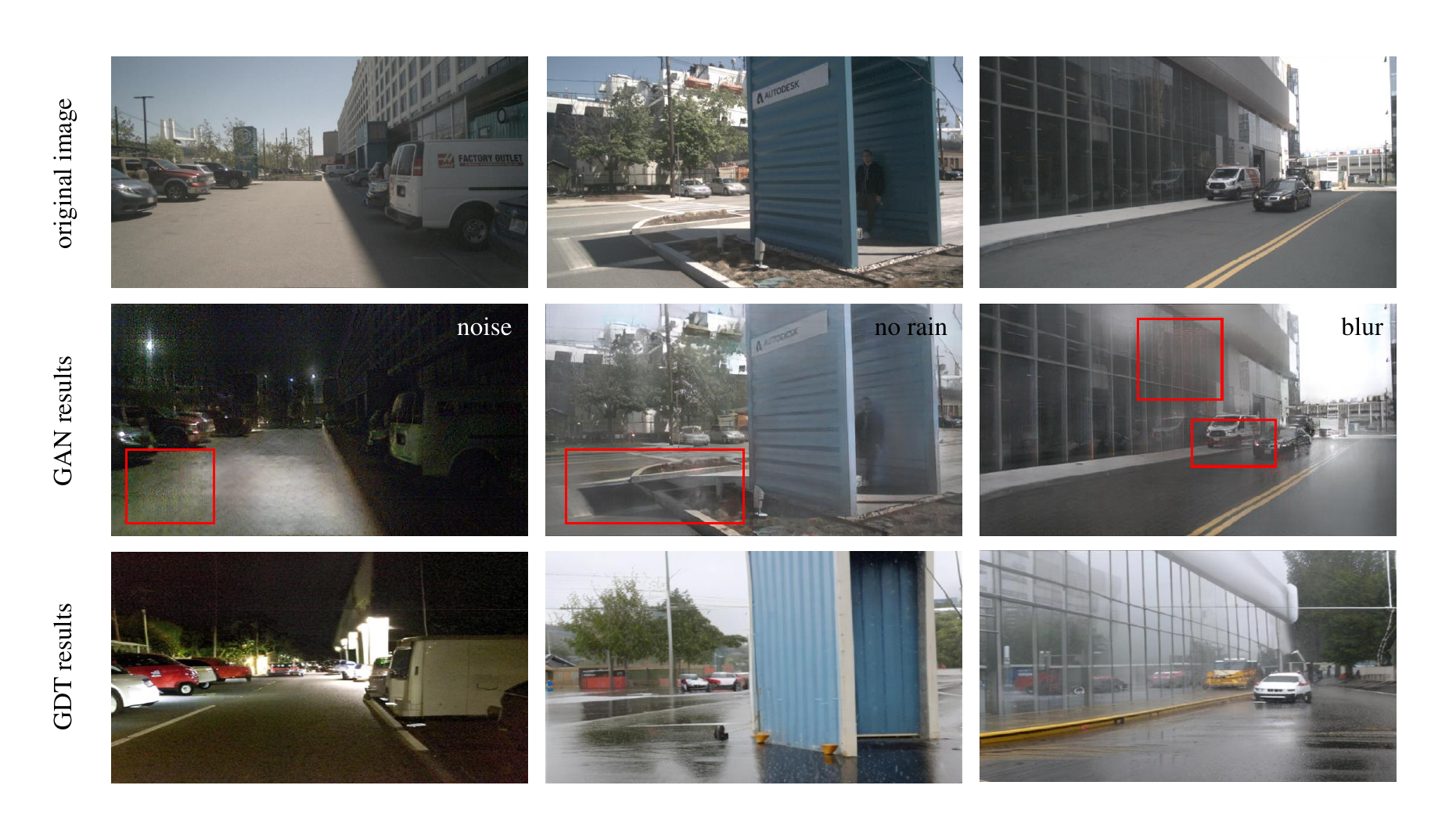}
\caption{\textbf{Shortcomings of GAN.} Compared to GAN, which suffers from issues like noise, fake rainy effects, and blurriness, our GDT can generate more diverse and realistic images.}
\label{fig:shortcommings_of_gan}
\vspace{-5mm}
\end{figure}

The main objective of this study is to introduce a comprehensive paradigm for Robust Monocular Depth Estimation (RMDE) aimed at overcoming the earlier-mentioned limitations. We aim to leverage valuable prior knowledge from stable diffusion \cite{Rombach_2022_CVPR} as the cornerstone of our approach. Modern generative diffusion models \cite{Rombach_2022_CVPR} have undergone extensive training on large-scale datasets, empowering them to produce high-quality images. Despite their potential, the use of generative diffusion models for RMDE remains largely unexplored in current literature. Figure \ref{fig:shortcommings_of_gan} visually illustrates the drawbacks of GAN-based translation models compared to our Generative Diffusion Model-based Translation (GDT) model. To address this gap, we employ a self-training approach that integrates the stable diffusion prior effectively. Our experiments confirm the feasibility and promising performance of this approach.

In our experiments, we found that the conventional ResNet architecture \cite{he2016deep} was not fully leveraging the potential of the stable diffusion prior. To overcome this limitation, we aimed to introduce a more potent feature encoder. Inspired by the Depth Anything approach \cite{depthanything}, we integrated DINOv2 into our encoder architecture to extract more effective and generic features from the samples, thereby enhancing the overall model performance. Furthermore, we noted the utility of semantic feature alignment in our work, prompting us to introduce semantic loss. Moreover, we introduced the teacher loss to improve the distillation process for depth estimation. This novel loss function facilitated the student model in acquiring the correct knowledge from the teacher model while preventing any erroneous knowledge transfer. Our contributions can be summarized as follows:

\begin{itemize}
\item We are the first to introduce stable diffusion into RMDE and propose a general paradigm that leverages the diffusion prior for robust depth estimation.

\item We present a plug-and-play translation model based on generative diffusion models that can be readily applied in various scenarios.

\item Our method outperforms existing approaches on the nuScenes and RobotCar datasets, achieving SOTA performance.
\end{itemize}

\section{Related Works}

\subsection{Monocular Depth Estimation}

\begin{figure}[t]
	\begin{center}
		\includegraphics[width=0.95\linewidth]{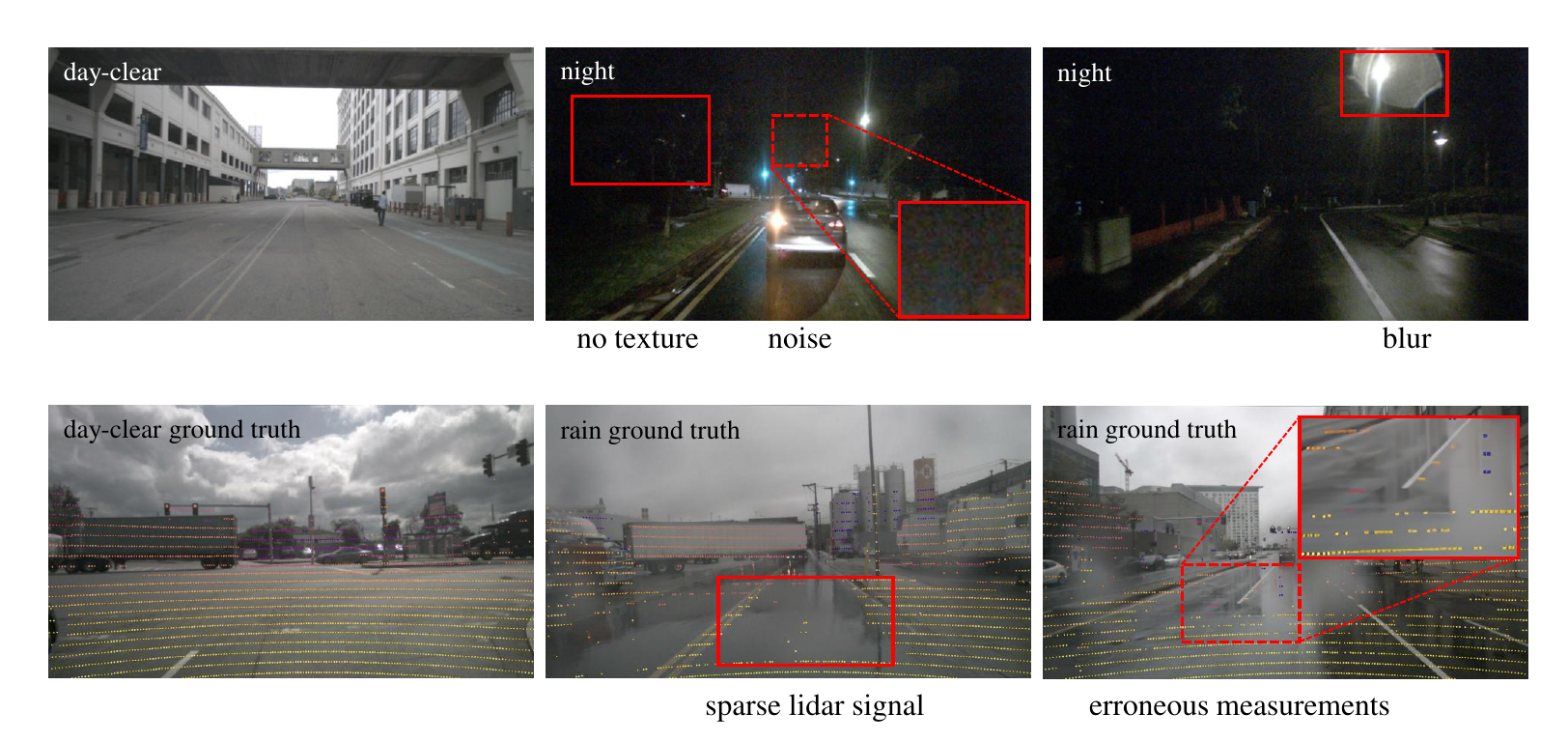}
	\end{center}
	\vspace{-0.7cm}
    \caption{\textbf{Darkness and weather effects on sensors.} The images above are from the nuScenes\cite{caesar2020nuscenes}. In night-time photos, RGB images often exhibit noise, textureless regions, and blurriness, which are not conducive to self-supervised learning. Additionally, rainy weather can introduce blur and reflections, leading to sparse and unreliable LiDAR signals, which are not suitable for supervised learning.}
	\label{fig:weather_affect}
 \vspace{-5mm}
\end{figure}

MDE methods can be categorized into two types: supervised and self-supervised. Eigen \textit{et al}. \cite{NIPS2014_multiscale_depth} introduced the CNN-based architecture for depth estimation, laying the foundation for supervised MDE. Over time, supervised MDE methods have evolved into regression methods\cite{Eigen_2015_ICCV, laina2016deeper} and classification methods\cite{fu2018deep, Adabins_Bhat_2021_CVPR, li2022binsformer}.
In contrast, self-supervised MDE does not rely on costly depth ground truth. It generates the supervisory signal using stereo image pairs\cite{10.1007/978-3-319-46484-8_45,Godard_2017_CVPR} or adjacent frames in a video\cite{Zhou_2017_CVPR,monodepth2}. These methods reconstruct the image based on the positional relationship between the cameras and the estimated depth, improving depth estimation accuracy by reducing the difference between the reconstructed image and the target image.
However, both supervised and self-supervised methods are susceptible to the effects of darkness and adverse weather conditions, as depicted in Figure \ref{fig:weather_affect}. Therefore, there is a need for RMDE methods capable of overcoming the poor performance of current approaches in challenging lighting conditions.

\subsection{Robust Monocular Depth Estimation}
In recent years, significant progress has been made in Robust Monocular Depth Estimation (RMDE), with several methods demonstrating promising results. These methods can be broadly categorized into two groups: model-based and data-based approaches.

Model-based methods aim to enhance the model architecture to handle challenging conditions.
DeFeat-Net \cite{spencer2020defeat} proposes a unified framework for learning robust monocular depth estimation and dense feature representation, specifically targeting improved performance under darkness.
RNW \cite{Wang_2021_ICCV} employs image enhancement techniques and an adversarial approach to enhance model performance in dark environments.
WSGD \cite{wsgd} addresses darkness by estimating flow maps and modeling light changes between adjacent frames.
MonoViT \cite{zhao2022monovit} utilizes Vision Transformer (ViT) \cite{dosovitskiy2020image} to extract image features, enabling the model to perform well under various weather conditions.

Data-based methods focus on leveraging additional modalities or augmenting the image signal through techniques such as domain adaptation.
DEISR \cite{9340998} utilizes sparse Radar data to enable depth estimation under adverse conditions.
R4Dyn \cite{9665968} demonstrates the benefits of weak supervision using sparse Radar data during training and the advantages of using Radar data as an additional input during inference for RMDE.
DET \cite{Shin_2023_CVPR} leverages thermal images to achieve depth estimation in darkness and adverse weather conditions.
ADFA \cite{adfa} employs adversarial training to enable the model to adapt to darkness.
ADDS \cite{liu2021self} utilizes different feature extractors to extract invariant and private features in different domains, enabling the extraction of universal image features and improving depth estimation.
ITDFA \cite{9803821} employs CycleGAN to translate images from daytime to other domains and extracts features from different domains to handle darkness and adverse weather.
Robust-Depth \cite{Saunders_2023_ICCV} introduces bi-directional pseudo-supervision loss and pseudo-supervised pose loss to compensate for the performance degradation caused by the use of translated images.
WeatherDepth \cite{wang2023weatherdepth} introduces contrastive learning based on domain adaptation methods.
Md4all \cite{md4all} achieves robust depth estimation by not distinguishing images in standard and challenging conditions, allowing the model to learn from the trained teacher network.

It is worth noting that the robustness of a model encompasses not only its ability to handle challenging conditions but also its generalization to other corruption types such as noise, blur, and digital artifacts. Several works \cite{ranftl2020towards, yin2021virtual, ranftl2021vision, birkl2023midas, xian2024towards} have explored the generalization of MDE models and strategies to improve their performance in various corruption types. In contrast, our work specifically focuses on enhancing the performance of MDE models in real-world weather corruptions and darkness.

\subsection{Generative Diffusion Models}

DDPM is a generative model, also known as a diffusion model, that achieves image generation by performing a diffusion process in the image space.
The impressive generative power of the diffusion model has led to the desire to incorporate control conditions into the generated images. In the field of text-based image generation, Rombach et al. proposed the latent diffusion model (LDM) \cite{Rombach_2022_CVPR}, which performs the diffusion process in the latent space. They also utilize the cross-attention mechanism \cite{vaswani2017attention} to introduce conditions for the LDM. Their text-to-image model is now known as Stable Diffusion.
To control the spatial structure of an image, Zhang et al. proposed ControlNet \cite{Zhang_2023_ICCV}, which provides an additional control image, such as a depth map, semantic map, canny map, etc., to govern the spatial structure of the generated image.
To control the style of generated images, Hu et al. proposed IP-Adapter \cite{ye2023ip-adapter}, an effective and lightweight neural network architecture aimed at achieving image prompt capability for pre-trained text-to-image diffusion models. The text-to-image diffusion models can generate images in the style of the images inputted into the IP-Adapter. IP-Adapter achieves this through a decoupled cross-attention mechanism where the cross-attention layers for text features and image features are separate. Our GDT model supports depth maps, text, and prompts for black-and-white or rainy day images as conditions, enabling the generation of images that satisfy the aforementioned conditions.

\section{Methodology}
In this paper, we proposed the SSD, a novel approach aims at stealing stable diffusion prior for RMDE. SSD incorporates a new translation model called GDT, which is based on generative diffusion models. To adapt to GDT for RMDE, we integrate DINOv2 into our depth model's architecture, it helps to extract universal image features. Besides, we optimized the distillation loss used for knowledge distillation. Our approach is generalized and can be adapted to a variety of challenging conditions.

\subsection{Preliminaries}
\label{Sec3.1}
In supervised monocular depth estimation, DepthNet are trained using sensor data as ground truth. The prediction process can be represented as $D = \mathcal{D}(I)$, where $I$ is the input image, $D$ is the dense depth map of $I$, and $\mathcal{D}$ is the DepthNet. For self-supervised MDE, adjacent frames in videos are used to train DepthNet. In addition to the DepthNet, a PoseNet is required to estimate the ego-motion between consecutive frames. For a frame $I_{t}$ in a video, DepthNet is used to predict the corresponding depth map $D_{t}$, and PoseNet is used to predict the camera ego-motion $P_{t \rightarrow t'}$. The process of predicting the camera ego-motion can be expressed as $P_{t \rightarrow t'} = \mathcal{P}(I_{t}, I_{t'})$, where $I_{t'}$ represents the adjacent frame of $I_{t}$ sampled from $\lbrace I_{t-1}, I_{t+1} \rbrace$, and $\mathcal{P}$ represents the PoseNet. The depth map and camera ego-motion are then used to synthesize the target image $I_t$. This process is described in Equation \ref{eqn:syn_img}, where $K$ represents the camera intrinsics, $Proj(\cdot)$ is the function that outputs the 2D coordinates, and $\langle \cdot \rangle$ is the sampling operator.
\begin{equation}
I_{t' \rightarrow t} = I_{t'}\langle Proj(D_{t}, P_{t \rightarrow t'}, K) \rangle
\label{eqn:syn_img}
\end{equation}

Finally, the photometric loss ($pe$) is calculated as defined in Equation \ref{eqn:photometric_loss}, where $\alpha$ is set to 0.85. In Monodepth2 \cite{monodepth2}, the per-pixel photometric loss $\mathcal{L}{p}$ is defined as $\min\limits{t'} pe(I_{t}, I_{t' \rightarrow t})$.

\begin{equation}
pe(I_{a}, I_{b}) = \frac{\alpha}{2}(1-SSIM(I_{a}, I_{b})) + (1 - \alpha)||I_{a} - I_{b}||
\label{eqn:photometric_loss}
\end{equation}

\subsection{Generative Diffusion Model-based Translation}
\label{Sec3.2}

\begin{figure}[h]
\centering
\includegraphics[width=0.9\linewidth]{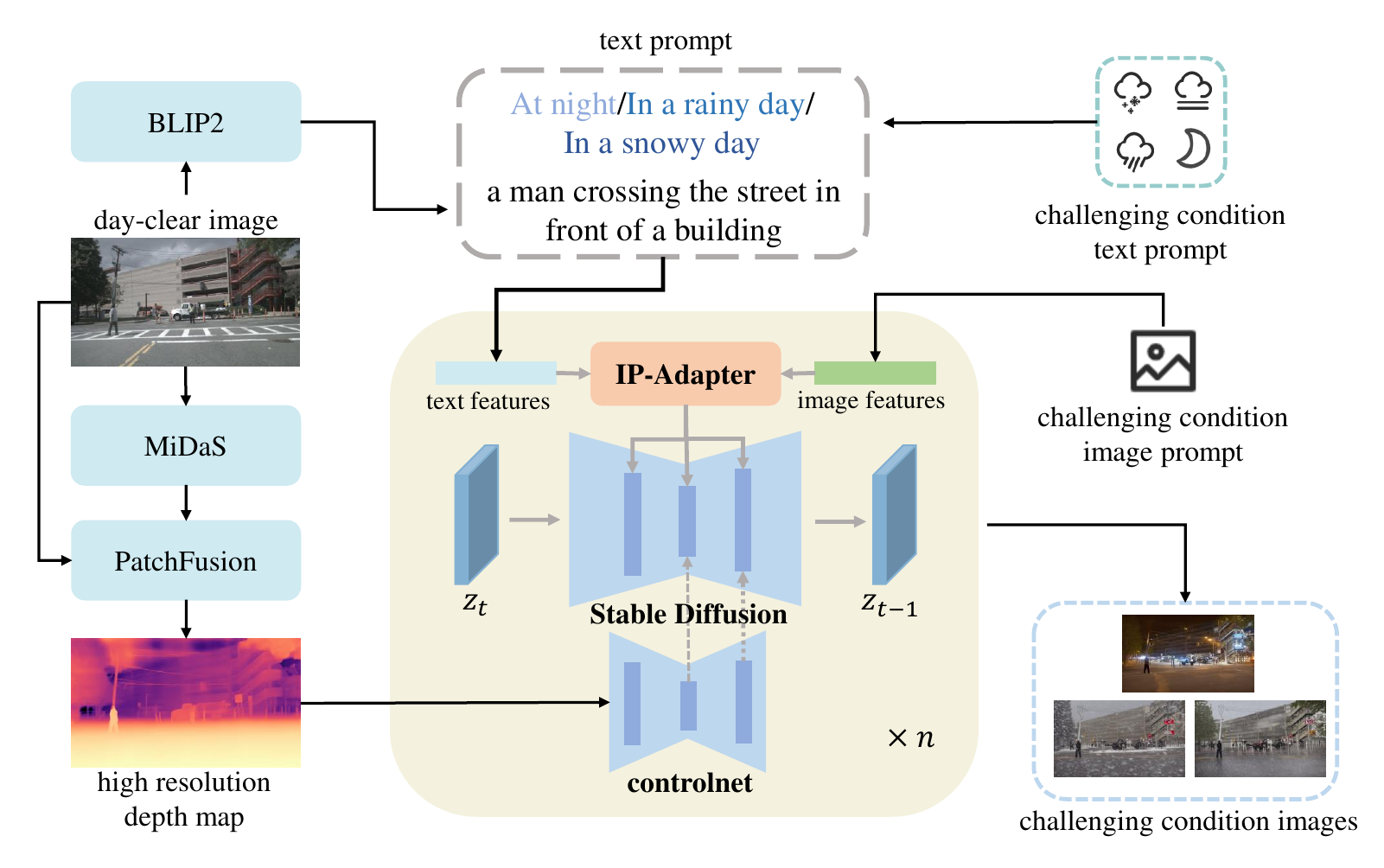}
\caption{The GDT pipeline incorporates multiple large models and PatchFusion to generate high-quality training samples.}
\label{fig:gdt_pipeline}
\vspace{-5mm}
\end{figure}

As discussed in Section \ref{sec:intro}, GAN-based translation models have exhibited certain limitations. In this paper, we present a novel translation model called GDT (Generative Diffusion Model-based Translation). The pipeline of GDT is depicted in Figure \ref{fig:gdt_pipeline}. The objective of GDT is to generate a training sample $I_g$ that closely resembles the day-clear image $I_d$ in terms of depth. To achieve this, we leverage the Stable Diffusion prior and introduce several control mechanisms to transform the day-clear image into challenging conditions while preserving specific characteristics.

First, we employ the BLIP2 model \cite{li2023blip} to obtain a scene description $T_{cap}$, which aids in preserving the content of the image. Second, we utilize the d2i model of the controlNet \cite{Zhang_2023_ICCV} to maintain approximate depth consistency. We have observed that the clearer the depth information, the more realistic the generated RGB image becomes. Thus, we employ PatchFusion to enhance the depth estimation obtained from the MiDaS model. Third, we address the challenge of introducing diverse styles into the generated images. We combine the $T_{cap}$ with challenging condition descriptors. However, we encountered an issue where the generated challenging condition images lacked realism. To overcome this, we introduce an image prompt for challenging conditions. In our implementation, we randomly select night images or rain images from the train dataset as image prompts. We utilize the IPAdapter model \cite{ye2023ip-adapter} to incorporate both text prompts and image prompts as inputs for image generation. The text prompt assists in preserving the content, while the image prompt facilitates style transfer.

In comparison to previous GAN-based methods, our GDT approach generates images with greater diversity and does not necessitate training for each specific scene. It can be considered a plug-and-play module. As more powerful generative diffusion models continue to emerge, our GDT method will become even more robust. The generation of the image $I_g$ can be expressed as follows:

\begin{equation}
\label{eqn:generate_image}
I_g = SD(IP(T_p, I_p),CN(D_h),z)
\end{equation}

Here, $SD$ represents the Stable Diffusion prior, $IP$ denotes the IPAdapter model that combines the text prompt $T_p$ and image prompt $I_p$, $CN$ corresponds to the controlNet model for depth consistency (utilizing $D_h$ as input), and $z$ represents the latent noise variable.

\begin{figure}[htbp!]
    \centering
    \includegraphics[width=0.9\linewidth]{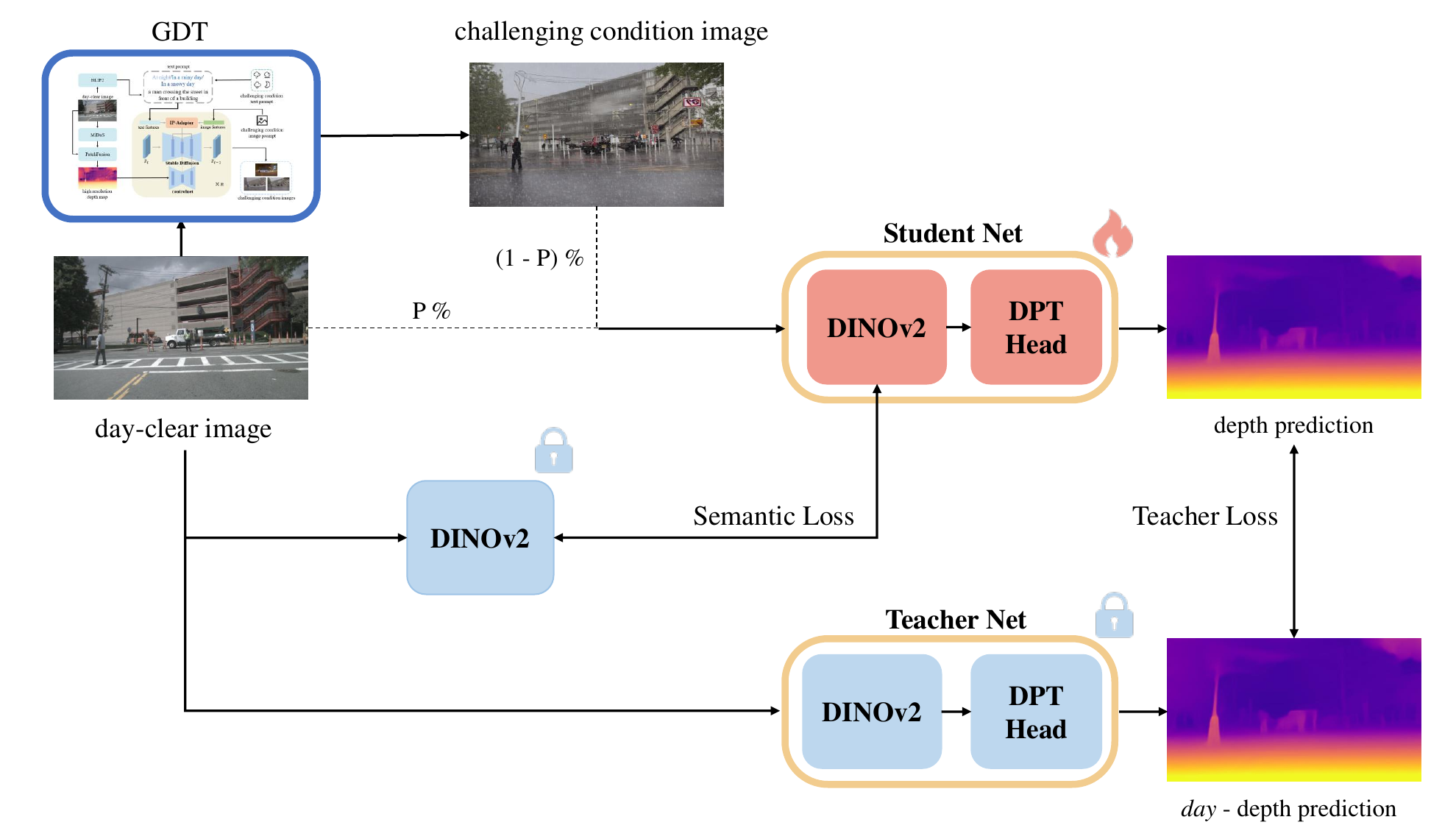}
    \caption{\textbf{SSD framework for robust depth estimation.} The Student Net receives guidance from the Teacher Net, leveraging a stable diffusion prior. The semantic loss ensures semantic consistency, while the teacher loss enables the Student Net to learn beyond the capabilities of the Teacher Net.}
    \label{fig:ssd_pipeline}
\vspace{-5mm}
\end{figure}

\subsection{Stealing Stable Diffusion Prior}
\label{Sec3.3}

In order to steal a stable diffusion prior for robust depth estimation, we have made modifications to the architecture of Monodepth2 \cite{monodepth2}. Additionally, we have employed a self-training strategy and introduced additional loss functions. Figure \ref{fig:ssd_pipeline} provides a brief illustration of our SSD pipeline.

\vspace{-3mm}
\subsubsection{Network Architecture}

In the context of generative diffusion model-based translation and GAN-based translation, the fundamental difference between them lies in their ability to generate data. Both approaches need to adhere to the assumption of constant depth before and after image translation. As a result, the translation process of GAN is more akin to style transfer, where the images before and after translation essentially belong to the same dataset.

However, the generative diffusion model-based translation, leveraging its powerful generative capabilities, guarantees that the image before and after translation will have the same depth while experiencing significant changes in content. This process can be regarded as a form of data scale-up. To achieve effective data scale-up, a simple encoder alone is insufficient in adequately extracting image features to further enhance model performance. Therefore, we have made modifications to the depth model based on Monodepth2 \cite{monodepth2}. We have moved away from the original ResNet-based architecture \cite{he2016deep} and introduced DINOv2 \cite{oquab2023dinov2} as the encoder in our depth model. This choice of encoder facilitates the extraction of robust image features. Additionally, we utilize pre-trained weights, enabling the depth model to inherit rich semantic information and scene understanding capabilities. Furthermore, we employ DPT depth head \cite{ranftl2021vision} as the decoder to regress depth. As for PoseNet, we maintain the original ResNet-based network architecture without making any changes.

\vspace{-3mm}
\subsubsection{Self-Training Strategy}
\label{Sec3.3.2}

Our self-training strategy follows the approach introduced in  \cite{md4all}, which has proven to be simple and effective in improving the performance of depth models. We adopt this strategy, which involves training a teacher network on day-clear samples and subsequently training a student network.

The first step is to train the teacher network using day-clear samples. This ensures that the teacher network produces reliable results specifically for day-clear scenarios. For instance, in the case of self-supervised monocular depth estimation, training the depth model using night samples can be challenging due to the lack of texture and the inability to utilize correspondences between adjacent frames. Photometric loss may fail in such scenarios.

Next, we train the student network to have the ability to perceive depth information across a variety of scenarios. To achieve this, we train the student network using a diverse range of training samples. For a day-clear training sample $I_d$, it can potentially be translated into a challenging condition training sample $I_g$. Here, we make the basic assumption that the depth between $I_d$ and $I_g$ remains relatively consistent. There is $P$ \% probability of no changes and $1-P$ \% probability of changing into $I_g$ where $P = |C|/(|C| + 1)$ and $|C|$ stands for the number of interest challenging conditions $C$. We leverage the teacher model to generate the pseudo ground truth $D_t$ for $I_d$. This pseudo ground truth $D_t$ also serves as a guidance for the student model when processing the GDT-translated $I_g$.

We then input a mixture of $I_d$ and $I_g$ into the student model and generate the predicted depth map $D_s$. By computing the distillation loss using $D_t$ and $D_s$, our model becomes capable of estimating depth not only under day-clear conditions but also under challenging conditions. This self-training strategy enables the student network to learn from both the teacher network and the GDT-translated samples, improving its ability to estimate depth across a diverse range of scenarios.

\vspace{-3mm}
\subsubsection{Loss computation}
 


We ultimately employed two loss functions: the teacher loss $\mathcal{L}_t$ and the semantic alignment loss $\mathcal{L}_s$. The teacher loss addresses the limitations of the naive distillation loss, while the semantic alignment loss draws inspiration from \cite{depthanything} and aims to better utilize the semantic similarity between $I_d$ and $I_g$. $\mathcal{L}_s$ is suitable for both supervised and self-supervised learning and $\mathcal{L}_t$ is only suitable for self-supervised learning and replaced by supervised loss in supervised learning.  

In self-supervised MDE, the teacher model struggles to accurately estimate the depth of each pixel, while the student model may outperform the teacher model in specific pixels. Although the teacher model initially provides guidance to the student model during training, as training progresses, the student model becomes more capable of independently estimating depth. In some cases, the student model's estimates may be more accurate than those of the teacher model. However, due to the limitations of the distillation loss used, the student model relies solely on the teacher model's estimates, hindering further improvement in its performance. On this basis, we propose the \textbf{teacher loss}. The core idea of teacher loss is to mask pixels corresponding to unreasonable depths estimated by the teacher model when calculating distillation loss.

For a day-clear frame $I_t$, we use the DepthNet of teacher model and student model to generate the corresponding dense depth map $D_t^t$ and $D_t^s$. Although the inputs of student model may be translated into $I_t^g$, we use the adjacent frames $I_{t'}$ to reconstruct the image $I_t$. Then, we use the PoseNet of the teacher model to estimate the camera ego-motion $P^t_{t \rightarrow t'}$. We obtain the $I^s_{t' \rightarrow t}$ and $I^t_{t' \rightarrow t}$ using the $D_t^s$, $D_t^t$ and $P^t_{t \rightarrow t'}$ as shown in equation \ref{eqn:syn_img_stu_tea}. 

\begin{equation}
\begin{split}
\label{eqn:syn_img_stu_tea}
    I^s_{t' \rightarrow t} = I_{t'}\langle Proj(D_{t}^s, P_{t \rightarrow t'}^t, K) \rangle \\
    I^t_{t' \rightarrow t} = I_{t'}\langle Proj(D_{t}^t, P_{t \rightarrow t'}^t, K) \rangle
\end{split}
\end{equation}

For self-supervised MDE, the lower the photometric loss, the more reasonable the estimated depth. If $D_t^t$ is more reasonable than $D_t^s$, then the photometric loss calculated by $D_t^t$ (T-$pe$) is lower than $D_t^s$ (S-$pe$). Thus, we design the mask $M$ to select the pixels with higher S-$pe$ than T-$pe$ as shown in equation \ref{equ:mask}, where $\lbrack \ \rbrack$ represents the Iverson brackets.

\begin{equation}
\label{equ:mask}
M = [ \min_{t'} pe(I_t, I^t_{t \rightarrow t'}) > \min_{t'} pe(I_t, I^s_{t \rightarrow t'})]
\end{equation}

We use the mask $M$ to filter the pixels with unreasonable depth. Our teacher loss is shown in equation \ref{equ:tea_loss} where $\mathcal{L}_d$ represents the day distillation loss defined in md4all-DD and $\odot$ represents the pixel-wise product. 

\begin{equation}
\label{equ:tea_loss}
\mathcal{L}_t = M \odot \mathcal{L}_d
\end{equation}

Besides teacher loss, we also used the \textbf{semantic loss} proposed in\cite{depthanything}. As the images translated by our GDT remain the semantic information similar to origin day-clear image $I_d$, we want to use the pre-trained DINOv2 to produce auxiliary image features for semantic align.    

\begin{equation}
    \label{equ:semantic_loss}
    \mathcal{L}_s = 1 - \frac{1}{HW} \sum_{i=1}^{HW} \cos(f_i, f_{i'})
\end{equation}

\section{Experiments and Results}
\label{sec:experiment}
\subsection{Experimental Setup}
\subsubsection{Datasets and Metrics}
In accordance with md4all \cite{md4all}, we choose \textbf{nuScenes} \cite{caesar2020nuscenes} and \textbf{Oxford RobotCar} \cite{robotcar} as the datasets for our study. NuScenes is a comprehensive dataset for autonomous driving, offering abundant sensor data and precise annotations. It encompasses around 1,000 urban driving scenes, encompassing diverse weather conditions, traffic scenarios, and road types. For the RMDE task, we partitioned the dataset based on visibility. Ordinarily, the dataset contains over 34,000 samples, with 28,130 samples allocated for training and 6,019 samples for validation. However, we utilize only 15,120 day-clear samples for training, and the remaining 6,019 samples are further divided based on visibility (i.e., day-clear, night, day-rain). The depth range for testing is 0.1-80 meters. RobotCar is a dataset gathered in Oxford, UK. In our study, we exclusively utilize day and night scenes, excluding rainy scenarios. The dataset comprises 16,563 day training samples and 1,411 test samples (702 day and 709 night). The depth range for testing is 0.1-50 meters.

In our experiments, we employ commonly adopted metrics for depth estimation, namely absRel, RMSE, $\delta_1$, and sqRel. Specifically, we use the first three metrics for evaluating the model's performance on nuScenes, while all four metrics are utilized for assessing the model's performance on RobotCar.

\vspace{-2mm}
\subsubsection{Implementation Details}
Both our teacher and student depth models utilize the architecture described in Section \ref{Sec3.3}.
In our experiments, we employ the DINOv2 encoder based on ViT-base as the encoder for our depth model, and we utilize ResNet-18 as the pose net encoder.
Both the student and teacher networks were trained for 20 epochs. The image input size for the pose encoder is 576$\times$320, while the DINOv2 encoder has an input size of 784$\times$518 (the output size is resized to 576$\times$320 for alignment).
During both stages of training, the learning rate for our depth model, which includes the DINOv2 encoder and DPT head, was set to 5e-6. The learning rate for the teacher's pose net was set to 5e-5 (the student's pose net was dropped).
We utilized the AdamW optimizer and applied a linear schedule to decay the learning rate. The batch size was set to 4. The parameters of the DINOv2 encoder were initialized with pre-trained weights.
For self-training, the DINOv2 encoder was guided by semantic loss using pre-trained weights, not the teacher DINOv2 encoder.
For GDT (Guided Depth Training), we utilized pre-trained models, including MiDaS, PatchFusion, BLIP-2, Stable Diffusion v1.5, ControlNet v1.1, and IP-Adapter, without further training.
Additionally, we chose the p49 mode for PatchFusion. Since the night images generated by SSD are brighter than typical night images, we adjusted their brightness to simulate more realistic outdoor night scenes.
All experiments were conducted on a single 24GB RTX 4090 GPU.

\subsection{Quantitative Results}

\begin{table}[t]
    \centering
    \label{nuScenes_results}
    \resizebox{\textwidth}{!}{
    \begin{tabular}{l@{\ }l@{\ }l|c c c|c c c|c c c}
        \toprule
        \multicolumn{3}{c|}{} & \multicolumn{3}{c|}{day-clear} & \multicolumn{3}{c|}{night} & \multicolumn{3}{c}{day-rain} \\
        Method & sup. & tr.data & absRel$ \,\downarrow$ & RMSE$\,\downarrow$ & $\delta_{1}\uparrow$ & absRel$ \,\downarrow$ & RMSE$ \,\downarrow$ & $\delta_{1}\uparrow$ & absRel$ \,\downarrow$ & RMSE$ \,\downarrow$ & $\delta_{1}\uparrow$ \\
        \midrule

        (a) MonoDepth2\cite{monodepth2} & M* & a & 0.1477 & 6.771 & 85.25 & 2.3332 & 32.940 & 10.54 & 0.4114 & 9.442 & 60.58 \\ 
            
        (b) PackNet-Sfm\cite{guizilini20203d} & Mv & d & 0.1567 & 7.230 & 82.64 & 0.2617 & 11.063 & 56.64 & 0.1645 & 8.288 & 77.07 \\

        (c) RNW\cite{Wang_2021_ICCV} & M* & dn & 0.2872 & 9.185 & 56.21 & 0.3333 & 10.098 & 43.72 & 0.2952 & 9.341 & 57.21 \\

        (d) md4all-baseline & Mv & d & 0.1333 & 6.459 & 85.88 & 0.2419 & 10.922 & 58.17 & 0.1572 & 7.453 & 79.49 \\

        (e) md4all-AD\cite{md4all} & Mv & dT(nr) & 0.1523 & 6.853 & 83.11 & 0.2187 & 9.003 & 68.84 &  0.1601 & 7.832 & 78.97 \\

        (f) md4all-DD\cite{md4all} & Mv & dT(nr) &  0.1366 & 6.452 & 84.61 & \pmb{0.1921} & 8.507 & \underline{71.07} & \underline{0.1414} & 7.228 & \underline{80.98} \\

        (g) SSD-T & M*v & d & \underline{0.1223} & \underline{6.132} & \pmb{87.39}& 0.2103& \pmb{8.024}& 67.70& 0.1718& \underline{6.988} & 78.87\\ 

        (h) SSD-S & M*vst & dT(nr) & \pmb{0.1217} & \pmb{5.982} & \underline{87.13} & \underline{0.1939} & \underline{8.038} & \pmb{72.31} & \pmb{0.1410} & \pmb{6.474} & \pmb{82.87}  \\

        \bottomrule
    \end{tabular}}

    \caption{Evaluation of self-supervised methods on the nuScenes\cite{caesar2020nuscenes} validation set. SSD-T: teacher model, SSD-S: student model, Supervisions(sup.): M: via monocular videos, *: test-time median-scaling via LiDAR, v: weak velocity loss, s: semantic loss, t: teacher loss. Training data (tr.data): d: day-clear, T: translated in, n: night, r: day-rain, a: all. Visual support: $\textbf{1}^{st}$, $\underline{2^{nd}}$.}
    \label{tab:quantitative_nuscenes}
    ~\vspace{-1cm}
\end{table}

The nuScenes dataset~\cite{caesar2020nuscenes} is divided into three scenarios: day-clear, night, and day-rain. Following the setup proposed in \cite{md4all}, we evaluated the performance of our model in each of these scenarios. The quantitative results are presented in Table \ref{tab:quantitative_nuscenes}.

Due to limitations imposed by the model architecture (which requires 3D convolutions) and the number of samples in the dataset, (b)PackNet's performance shows slight improvement for night and rain scenes but is greatly reduced for day-clear scenes. The weak velocity loss in (b)PackNet contributes to obtaining metric depth. Although (c)RNW employs a unique image enhancement method and adversarial training, it performs worse than our baseline. On the other hand, (e)md4all-AD and (f)md4all-DD are simple and efficient methods that enhance the model's robustness. Building upon the md4all approach, we have made improvements to the model architecture and translation model.
Comparing (d) and (g), our model architecture (DINOv2 encoder with DPT head) demonstrates significant superiority over monoDepth2. Additionally, thanks to the pre-trained weights from the DINOv2 encoder, our depth model can extract more robust features.
Comparing (g) and (h), our student model outperforms the teacher model. In the night scenario, absRel is reduced by 7.8\%, and in the day-rain scenario, absRel is reduced by 17.9\%. This demonstrates that the student model indeed benefits from the GDT-transformed images and effectively incorporates the stable diffusion prior.

\begin{table*}
\vspace{-5mm}
\begin{center}
\resizebox{\textwidth}{!}{
\begin{tabular}{lll|cccc|cccc}
    \toprule
    &&& \multicolumn{4}{c|}{\textit{day} -- RobotCar} & \multicolumn{4}{c}{\textit{night} -- RobotCar} \\
    Method  & sup. & tr.data & absRel & sqRel & RMSE & $\delta_1$   & absRel & sqRel & RMSE & $\delta_1$   \\
    \midrule

    (a) Monodepth2~\cite{monodepth2}   & M$^*$ & d & 0.1196 & 0.670 & 3.164 & 86.38 & 0.3029 & 1.724 & 5.038 & 45.88 \\
    (b) DeFeatNet~\cite{spencer2020defeat}  & M$^*$ & a & 0.2470 & 2.980 & 7.884 & 65.00 & 0.3340 & 4.589 & 8.606 & 58.60\\
    (c) ADIDS~\cite{Liu_2021_ICCV}  & M$^*$ & a & 0.2390 & 2.089 & 6.743 & 61.40 & 0.2870 & 2.569 & 7.985 & 49.00 \\
    (d) RNW~\cite{Wang_2021_ICCV}   & M$^*$ & a & 0.2970 & 2.608 & 7.996 & 43.10 & 0.1850 & 1.710 & 6.549 & 73.30 \\
    (e) WSGD~\cite{wsgd}   & M$^*$ & a & 0.1760 & 1.603 & 6.036 & 75.00 & 0.1740 & 1.637 & 6.302 & 75.40 \\
    (f) md4all-DD~\cite{md4all}  & Mv & dT(n) & 0.1128 & 0.648 & 3.206 & 87.13 & 0.1219 & \underline{0.784} & \underline{3.604} & 84.86 \\
    (g) SSD-T  & Mv & dT(n) & \underline{0.1069} & \underline{0.762} & \underline{3.241} & \pmb{89.38} & \pmb{0.1137} & \pmb{0.685} & \pmb{3.460} & \underline{86.14} \\
    (h) SSD-S  & Mv & dT(n) & \pmb{0.1058}& \pmb{0.718}& \pmb{3.203} & \underline{89.09} & \underline{0.1143}& 0.823& 3.646 & \pmb{87.36} \\

\bottomrule
\end{tabular}
}

\end{center}
\vspace{-0.2cm}
\caption{Evaluation of self-supervised works on the RobotCar~\cite{robotcar} test set. Notation from Table~\ref{tab:quantitative_nuscenes}.}
\label{table:quantitative_robotcar}
\vspace{-1cm}
\end{table*}

The RobotCar dataset~\cite{robotcar} is divided into day and night scenarios. (a), (f), and (g) demonstrate the strong generalization capability of our model architecture. Despite not being trained on night data, our model performs well and even outperforms md4all-DD. Furthermore, (g) and (h) illustrate that, despite already achieving high performance, our GDT provides a prior that enables the student model to surpass the teacher model. This leads to further improvements in depth estimation accuracy for both day and night scenarios.

\subsection{Qualitative Results}

\begin{figure}[t]
\begin{center}
\includegraphics[width=0.97\textwidth]{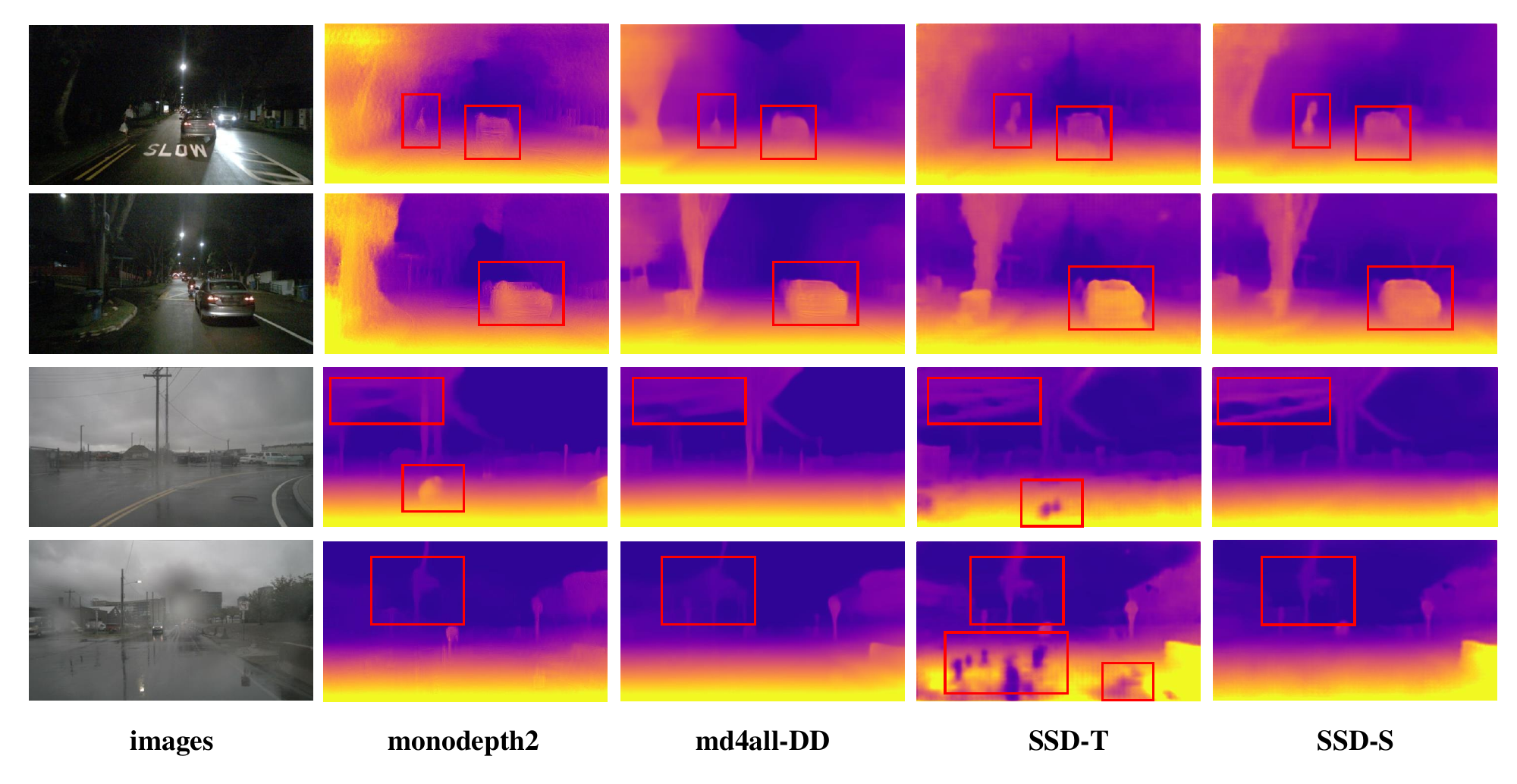}
\end{center}
\vspace{-0.5cm}
\caption{Comparison of samples from the nuScenes dataset~\cite{caesar2020nuscenes} among monodepth2~\cite{monodepth2}, md4all-DD~\cite{md4all}, and our self-supervised teacher model SSD-T, as well as the student model SSD-S.}
\label{fig:qualitative_nuscenes_comp}
\end{figure}

\begin{figure}[t]
\vspace{-0.2cm}
\begin{center}
\includegraphics[width=1.00\linewidth]{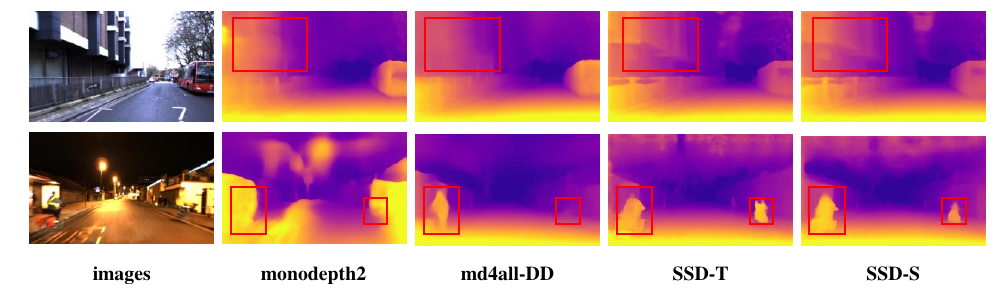}
\end{center}
\vspace{-0.5cm}
\caption{Comparison of samples from the RobotCar dataset~\cite{robotcar} among monodepth2~\cite{monodepth2}, md4all-DD~\cite{md4all}, and our self-supervised teacher model SSD-T, as well as the student model SSD-S.}
\label{fig:qualitative_robotcar_comp}
\vspace{-0.55cm}
\end{figure}

The qualitative results depicted in Figure \ref{fig:qualitative_nuscenes_comp} and Figure \ref{fig:qualitative_robotcar_comp} showcase the performance of our SSD (Single Shot Depth) model in depth estimation, specifically on the nuScenes~\cite{caesar2020nuscenes} and RobotCar~\cite{robotcar} datasets. In the nuScenes dataset, our SSD model demonstrates enhanced performance in extracting depth information from regions with low illumination, such as night scenes. Moreover, our approach effectively restores depth information even in rainy scenes. Figure \ref{fig:qualitative_nuscenes_comp} highlights the advantages of our SSD-T (teacher) and SSD-S (student) models over the monodepth2 and md4all-DD methods in night scenes. Our models accurately estimate the depth of cars and pedestrians, whereas the other methods fail to do so. However, in rainy scenes, SSD-T's performance is compromised due to the superior perceptive ability of the DINOv2 architecture. SSD-T mistakenly perceives virtual images in water as objects, leading to incorrect depth estimations. Nonetheless, SSD-S, which incorporates semantic loss and teacher loss, achieves excellent results. Figure \ref{fig:qualitative_robotcar_comp} displays the results obtained in both day-clear and night scenes. In the day-clear scene, both SSD-T and SSD-S successfully capture the edge of the wall in their depth maps, whereas monodepth2 and md4all-DD fail to do so. In the night scene, monodepth2 produces unreliable depth maps, whereas our SSD-T and SSD-S accurately estimate the depth of the bicycle on the right side.

\subsection{Ablation Study}

We conducted ablation experiments on the nuScenes validation set to validate the effectiveness of our proposed approach. The results are presented in Table \ref{tab:ablation_nuscenes}. In the ablation study, our metrics represent the average performance of the model across all samples in the validation set. 
In Method 1, we changed the original translation model of md4all to our GDT, without incorporating PatchFusion. The performance of the model improved as GDT can produce detailed images that preserve fine depth information.
In Method 2, we introduced PatchFusion, which enhanced the model's performance by generating high-resolution depth maps. These maps contribute to the preservation of fine depth details and aid GDT in producing more accurate images.
During knowledge distillation, the teacher model may produce unreasonable depth estimations. In Method 3, we replaced the distillation loss with the teacher loss, resolving this issue and further improving the model's performance.
Considering the powerful generative capability of SSD, it is necessary to have a more robust backbone to produce reliable image features. DINOv2, known for its ability to generate robust image features, is suitable for our task. In Method 4, after adopting DINOv2 as our encoder, the performance of the depth model improved significantly.
In Method 5, we employed semantic loss to align the image features, resulting in a further improvement in model performance.

\begin{table}[h]
    \centering
    \resizebox{\textwidth}{!}{
        \begin{tabular}{ l | c | c | c | c | c | c | c | c| c | c | c}
        \toprule
        \multirow{2}{*}{method} & \multicolumn{2}{c|}{backbone} & \multicolumn{3}{c|}{translation model} & \multicolumn{3}{c|}{loss} & \multicolumn{3}{c}{metrics-\textit{all}} \\
         & ResNet-18 & DINOv2  & GAN & MiDaS & PatchFusion  & distillation loss & teacher loss & semantic loss & absRel$ \,\downarrow$ & RMSE$ \,\downarrow$ & $\delta_1\uparrow$ \\
        \midrule
    
       0 & \checkmark & & \checkmark & & & \checkmark & & & 0.1513 & 6.761 & 81.51 \\
        
       1 & \checkmark & & & \checkmark & & \checkmark & & & 0.1529 & 6.887 & 81.03 \\

       2 & \checkmark & & & \checkmark & \checkmark & \checkmark & & & 0.1497 & 6.806 & 81.16 \\

       3 & \checkmark & & & \checkmark & \checkmark & &\checkmark & & 0.1485& 6.801& 81.30\\

       4 & & \checkmark & & \checkmark & \checkmark & & \checkmark & &0.1360 & 6.333& 84.07\\

       5 & & \checkmark &  & \checkmark & \checkmark & & \checkmark & \checkmark & 0.1320 & 6.266 & 84.96 \\

        \bottomrule 
        \end{tabular}
    }
    \caption{Ablation study on the nuScenes dataset. We compare the effects of different training settings on the depth model.}
    \label{tab:ablation_nuscenes}
    \vspace{-1cm}
\end{table}

\section{Conclusion}
This paper introduces SSD, a practical solution for robust monocular depth estimation. We underscore the significance of the generative diffusion model prior in generating challenging samples. We have successfully integrated the stable diffusion prior into our depth estimation model using a self-training approach. With the emergence of stronger generative diffusion models and more accurate control models, along with advancements in depth model capabilities, the potential for robust depth estimation can be further enhanced. Furthermore, the SSD paradigm can extend to other dense prediction tasks, such as robust semantic segmentation and 3D occupancy prediction, offering promising avenues for future research and applications.

%
%
\clearpage
%
%
\bibliographystyle{splncs04}
\bibliography{egbib}
\end{document}